\documentclass{article}
\usepackage{mlspconf}
\usepackage{amsmath,amssymb}
\usepackage[dvipdfmx]{graphicx}
\usepackage{circle}
\newcommand{\bm}[1]{\mathbf{#1}}

\def\sf#1{\textsf{#1}}
\def\rm#1{\textrm{#1}}
\def\tr{\rm{tr}}

\def\Nc#1#2#3{\mathcal{N}_\mathbb{C}( #1 | #2, #3 )}
\def\N#1#2#3{\mathcal{N}( #1 | #2, #3 )}
\def\eqcite#1{Eq.~(\ref{eq:#1})}
\def\figcite#1{Fig.~\ref{fig:#1}}

\title{Generalized Multichannel Variational Autoencoder for Underdetermined source separation}
\name{
Shogo Seki$^{1)}$, 
Hirokazu Kameoka$^{2)}$,
Li Li$^{3)}$, 
Tomoki Toda$^{4)}$, 
Kazuya Takeda$^{5)}$}
\address{
$^{1)}$Graduate School of Informatics, Nagoya University, Japan\\
$^{2)}$NTT Communication Science Laboratories, Nippon Telegraph and Telephone Corporation, Japan\\
$^{3)}$Graduate School of Systems and Information Engineering, University of Tsukuba, Japan\\
$^{4)}$Information Technology Center, Nagoya University, Japan\\
$^{5)}$Institutes of Innovation for Future Society, Nagoya University, Japan
}
\begin{document}
\ninept

\maketitle
\begin{abstract}
	This paper deals with a multichannel audio source separation problem under underdetermined conditions.
	Multichannel Non-negative Matrix Factorization (MNMF) is one of powerful approaches, which adopts the NMF concept for source power spectrogram modeling.
	This concept is also employed in Independent Low-Rank Matrix Analysis (ILRMA), a special class of the MNMF framework formulated under determined conditions.
	While these methods work reasonably well for particular types of sound sources, one limitation is that they can fail to work for sources with spectrograms that do not comply with the NMF model.
	To address this limitation, an extension of ILRMA called the Multichannel Variational Autoencoder (MVAE) method was recently proposed, where a Conditional VAE (CVAE) is used instead of the NMF model for source power spectrogram modeling.
	This approach has shown to perform impressively in determined source separation tasks thanks to the representation power of DNNs.
	While the original MVAE method was formulated under determined mixing conditions, this paper generalizes it so that it can also deal with underdetermined cases.
	We call the proposed framework the Generalized MVAE (GMVAE).
	The proposed method was evaluated on a underdetermined source separation task of separating out three sources from two microphone inputs.
	Experimental results revealed that the GMVAE method achieved better performance than the MNMF method.

\end{abstract}
\begin{keywords}
Underdetermined source separation, Multichannel non-negative matrix factorization, Multichannel audoencoder
\end{keywords}
\section{Introduction}
	Blind source separation (BSS) refers to a problem of separating out individual source signals from microphone array inputs where the transfer functions between the sources and microphones are unknown.
	The frequency-domain BSS approach allows the utilization of various models for the time-frequency representations of source signals and/or array responses. For example, Independent Vector Analysis (IVA)~\cite{kim2006independent,hiroe2006solution} offers a way of jointly solving frequency-wise source separation and permutation alignment under the assumption that the magnitudes of the frequency components originating from the same source are likely to vary coherently over time.

	Other approaches involve multichannel extensions of Non-negative Matrix Factorization (NMF)~\cite{ozerov2010multichannel,kameoka2010statistical,sawada2013multichannel,kitamura2016determined,kitamura2018determined}.
	NMF was originally applied to music transcription and monaural source separation tasks~\cite{smaragdis2003non,fevotte2009nonnegative} where the idea is to interpret the power spectrogram of a mixture signal and approximate it as the product of two non-negative matrices.
	This can be viewed as approximating the power spectrum of a mixture signal observed at each time frame by the sum of basis spectra scaled by time-varying magnitudes.
	Multichannel NMF (MNMF) is an extension of this approach to a multichannel case that allows for the use of spatial information.
	It can also be seen as an approach to frequency-domain BSS using spectral templates as a clue for jointly solving frequency-wise source separation and permutation alignment.

	The original MNMF~\cite{ozerov2010multichannel} was formulated under a general problem setting where sources can outnumber microphones and a determined version of MNMF was subsequently proposed in~\cite{kameoka2010statistical}. 
	While the determined version is applicable only to determined cases, it allows an implementation of a significantly faster algorithm than the general version. 
	The determined MNMF framework was later called Independent Low-Rank Matrix Analysis (ILRMA)~\cite{kitamura2018determined}. 
	The MNMF framework including ILRMA is notable in that the optimization algorithm is guaranteed to converge, however, one limitation is that it can fail to work for sources with spectrograms that do not comply with the NMF model.
	
	To address this limitation, a technique called the Multichannel Variational Autoencoder (MVAE) method was recently proposed in~\cite{kameoka2018semi}.
	It is an extension of ILRMA in which a Conditional VAE (CVAE)~\cite{kingma2014semi} is used instead of the NMF model to estimate the power spectrograms of the sources in a mixture.
	Specifically, MVAE allows the estimation of the separation matrices by employing a single CVAE, trained using the spectrograms of speech samples with speaker ID labels, as a generative model of the speech spectrograms of multiple speakers.
	This approach is noteworthy in that it can exploit the benefits of the representation power of DNNs for source power spectrogram modeling and has shown to outperform ILRMA on a determined source separation task.

	While the original MVAE method was formulated under determined mixing conditions, this paper generalizes it so that it can also deal with underdetermined cases.
	We call the proposed framework Generalized MVAE (GMVAE) to distinguish it from the original determined version.
	
\section{Problem Formulation}
We consider a situation where $J$ source signals are observed by $I$ microphones.
Let $s_j(f, n)$ and $x_i(f, n)$ be the Short-Time Fourier Transform (STFT) coefficient of the $j$-th source signal and the $i$-th observed signal, where $f$ and $n$ are the frequency and time indices, respectively.
We denote the vectors containing $s_1(f, n), \cdots, s_J(f, n)$ and $x_1(f, n), \cdots, x_I(f, n)$ by
\begin{align}
	\bm{s}(f, n) = \left[ s_1(f, n), \cdots, s_J(f, n) \right]^\sf{T} \in \mathbb{C}^J, \\
	\bm{x}(f, n) = \left[ x_1(f, n), \cdots, x_I(f, n) \right]^\sf{T} \in \mathbb{C}^I,
\end{align}
where $(\cdot)^\sf{T}$ denotes transpose.
Now, we use a mixing system of the form 
\begin{align}
	\label{eq:mixing}
	\bm{x}(f, n) &= \bm{A}(f)\bm{s}(f, n), \\
	\bm{A}(f) &= \left[ \bm{a}_1(f), \cdots, \bm{a}_J(f) \right] \in \mathbb{C}^{I \times J},
\end{align}
to describe the relationship between $\bm{s}(f, n)$ and $\bm{x}(f, n)$ where $\bm{A}(f)$ is called the mixing matrix.

Here, we assume that $s_j(f, n)$ independently follows a zero-mean complex Gaussian distribution with variance $v_j(f, n) = \mathbb{E}\left[ |s_j(f, n)|^2 \right]$
\begin{align}
	\label{eq:LGM}
	s_j(f, n) \sim \Nc{s_j(f, n)}{0}{v_j(f, n)}.
\end{align}
\eqcite{LGM} is called the Local Gaussian Model~(LGM).
When $s_j(f, n)$ and $s_{j'}(f, n)$ are independent for $j \neq j'$, $\bm{s}(f, n)$ follows
\begin{align}
	\label{eq:LGMs}
	\bm{s}(f, n) \sim \Nc{\bm{s}(f, n)}{\bm{0}}{\bm{V}(f, n)},
\end{align}
where $\bm{V}(f, n)$ is a diagonal matrix with diagonal entries $v_1(f, n)$, $\cdots$, $v_J(f, n)$.
From \eqcite{mixing} and \eqcite{LGMs}, $\bm{x}(f, n)$ is shown to follow
\begin{align}
	\label{eq:observed}
	\bm{x}(f, n) \sim \Nc{\bm{x}(f, n)}{\bm{0}}{\bm{A}(f)\bm{V}(f, n)\bm{A}^\sf{H}(f)},
\end{align}
where  $(\cdot)^\sf{H}$ denotes conjugate transpose.
Thus, the log-likelihood of the mixing matrices $\mathcal{A}=\left\{ \bm{A}(f) \right\}_f$ and the variances of source signals $\mathcal{V}=\left\{ v_j(f, n) \right\}_{f, n}$ given the observed mixture signals $\mathcal{X}=\left\{ \bm{x}(f, n) \right\}_{f, n}$ is given by
\begin{align}
	\label{eq:likelihood}
	&\log p(\mathcal{X}|\mathcal{A}, \mathcal{V})  \overset{c}{=} \nonumber\\
	&-\sum_{f, n}\left[\rm{tr}( \bm{x}^\sf{H}(f, n)(\bm{A}(f)\bm{V}(f, n)\bm{A}^\sf{H}(f))^{-1} \bm{x}(f, n)) \right. \nonumber\\
	&\left. \qquad\quad+ \rm{logdet} (\bm{A}(f)\bm{V}(f, n)\bm{A}^\sf{H}(f)) \right],
\end{align}
where $\overset{c}{=}$ denotes equality up to constant terms. 
If there is no constraint imposed on $v_j(f, n)$, \eqcite{likelihood} will be split into frequency-wise source separation problems.
This indicates that there is a permutation ambiguity in the separated components for each frequency since permutation of $j$ does not affect the value of the log-likelihood.
Thus, we usually need to perform permutation alignment after $\mathcal{A}$ is obtained.

\section{Related Work}
\subsection{MNMF}
The spatial covariance of the observed mixture signal $\bm{A}(f) \bm{V}(f, n) \bm{A}^\sf{H}(f, n)$ can be rewritten as the linear sum of the outer products of $\bm{a}_j(f)$ multiplied by $v_j(f, n)$:
\begin{align}
	\label{eq:mnmf}
	\bm{A}(f)\bm{V}(f, n)\bm{A}^\sf{H}(f) &= \sum_j\bm{a}_j(f)v_j(f, n)\bm{a}_j^\sf{H}(f) \nonumber\\
	&= \sum_jv_j(f, n)\bm{R}_j(f),
\end{align}
where $\bm{R}_j(f)$ represents the spatial covariance of source $j$.
As with IVA, MNMF makes it possible to jointly solve frequency-wise source separation and permutation alignment by imposing a constraint on $v_j(f, n)$.
Specifically, $v_j(f, n)$ is modeled as the linear sum of $K_j$ spectral templates $h_{j, 1}(f)$, $\cdots$, $h_{j, K_j}(f) \ge 0$ scaled by time-varying activations $u_{j, 1}(n), \cdots, u_{j, K_j}(n) \ge 0$:
\begin{align}
	\label{eq:nmf}
	v_j(f, n) = \sum_{k=1}^{K_j}h_{j, k}(f) u_{j, k}(n).
\end{align}
It is also possible to allow all the spectral templates to be shared by every source and let the contribution of the $k$-th spectral template to source $j$ be determined in a data-driven manner.
Thus, $v_j(f, n)$ can also be expressed as
\begin{align}
	\label{eq:nmf2}
	v_j(f, n) = \sum_{k=1}^{K} b_{j, k} h_{k}(f) u_{k}(n),
\end{align}
where $b_{j, k} \in [0, 1]$ is a continuous indicator variable that satisfies $\sum_kb_{j, k} = 1$.
Here, $b_{j, k}$ can be interpreted as the expectation of a binary indicator variable that describes the index of the source to which the $k$-th template is assigned.

The optimization algorithm of MNMF consists of iteratively updating the spatial covariance matrices $\mathcal{R} = \left\{ \bm{R}_j(f) \right\}_{j, f}$, and the source models $\mathcal{H}_1 = \left\{ h_{j, k}(f) \right\}_{j, k, f}$, $\mathcal{U}_1 = \left\{ u_{j, k}(n) \right\}_{j, k, n}$ or $\mathcal{B} = \left\{ b_{j, k} \right\}_{j, k}$, $\mathcal{H}_2 = \left\{ h_{k}(f) \right\}_{k, f}$, $\mathcal{U}_2 = \left\{ u_{k}(n) \right\}_{k, n}$.
We can derive update equations using the principle of the Majorization-Minimization (MM) algorithm.
For the update of $\mathcal{R}$, we can use the solution to Algebraic Ricatti equation
\begin{align}
	\label{eq:ricatti}
	\bm{R}_j(f)\bm{\Psi}_j(f)\bm{R}_j(f) &= \bm{\Omega}_j(f),
\end{align}
where
\begin{align}
	\label{eq:ricatti_1}
	\bm{\Psi}_j(f) &= \sum_{n} v_j(f, n)\hat{\sf{X}}^{-1}(f, n), \\
	\label{eq:ricatti_2}
	\bm{\Omega}_j(f) &= \nonumber \\
	\bm{R}_j(f)&\left( \sum_nv_j(f, n)\hat{\sf{X}}^{-1}(f, n)\sf{X}(f, n)\hat{\sf{X}}^{-1}(f, n) \right)\bm{R}_j(f).
\end{align}
Note that we have used $\sf{X}(f, n)$ and $\hat{\sf{X}}(f, n)$ to represent
\begin{align}
	\label{eq:relation1}
	\sf{X}(f, n) &= \bm{x}(f, n)\bm{x}^\sf{H}(f, n), \\
	\label{eq:relation2}
	\hat{\sf{X}}(f, n) &= \bm{A}_j(f)\bm{V}(f, n)\bm{A}_j(f).
\end{align}
Performing $\bm{R}_j(f) \leftarrow \left(\bm{R}_j(f) + \bm{R}_j^\sf{H}(f)\right) / \ 2$ and $\bm{R}_j(f) \leftarrow \bm{R}_j(f)  + \epsilon \bm{I}$ after solving \eqcite{ricatti} is empirically shown to help avoid computational instability.
As in~\cite{sawada2013multichannel}, update rules for $\mathcal{H}_1$ and $\mathcal{U}_1$ can be derived as 
\begin{align}
	&h_{j, k}(f) \leftarrow h_{j, k}(f)\nonumber\\
	&\times \sqrt{ \frac{ \sum_{n} u_{j, k}(n) \rm{tr}(\hat{\sf{X}}^{-1}(f, n)\sf{X}(f, n)\hat{\sf{X}}^{-1}(f, n)\bm{R}_j(f)) }{ \sum_{n} u_{j, k}(n) \rm{tr}(\hat{\sf{X}}^{-1}(f, n)\bm{R}_j(f)) } },\\
	&u_{j, k}(n) \leftarrow u_{j, k}(n)\nonumber\\
	&\times \sqrt{ \frac{ \sum_{f} h_{j, k}(f) \rm{tr}(\hat{\sf{X}}^{-1}(f, n)\sf{X}(f, n)\hat{\sf{X}}^{-1}(f, n)\bm{R}_j(f)) }{ \sum_{f} h_{j, k}(f) \rm{tr}(\hat{\sf{X}}^{-1}(f, n)\bm{R}_j(f)) } }.
\end{align}
When $v_j(f, n)$ is given in the form of \eqcite{nmf2}, update rules for $\mathcal{B}$, $\mathcal{H}_2$, and $\mathcal{U}_2$ can be derived as
\begin{align}
	\label{eq:b}
	&b_{j, k} \leftarrow b_{j, k}\nonumber\\
	&\times \sqrt{ \frac{ \sum_{f, n} h_{k}(f) u_{k}(n) \rm{tr}(\hat{\sf{X}}^{-1}(f, n)\sf{X}(f, n)\hat{\sf{X}}^{-1}(f, n)\bm{R}_j(f)) }{ \sum_{f, n} h_{k}(f) u_{k}(n)  \rm{tr}(\hat{\sf{X}}^{-1}(f, n)\bm{R}_j(f)) } },\\
	&h_{k}(f) \leftarrow h_{k}(f)\nonumber\\
	&\times \sqrt{ \frac{ \sum_{j, n} b_{j, k}u_{k}(n) \rm{tr}(\hat{\sf{X}}^{-1}(f, n)\sf{X}(f, n)\hat{\sf{X}}^{-1}(f, n)\bm{R}_j(f)) }{  \sum_{j, n} b_{j, k}u_{k}(n) \rm{tr}(\hat{\sf{X}}^{-1}(f, n)\bm{R}_j(f)) } },\\
	&u_{k}(n) \leftarrow u_{k}(n)\nonumber\\
	&\times \sqrt{ \frac{ \sum_{j, f} b_{j, k}h_{k}(f) \rm{tr}(\hat{\sf{X}}^{-1}(f, n)\sf{X}(f, n)\hat{\sf{X}}^{-1}(f, n)\bm{R}_j(f)) }{  \sum_{j, f} b_{j, k}h_{k}(f) \rm{tr}(\hat{\sf{X}}^{-1}(f, n)\bm{R}_j(f)) } }.
\end{align}
To ensure that $\mathcal{B}$  satisfies the sum-to-one constraint, we normalize $\mathcal{B}$ after performing \eqcite{b} by  $b_{j, k}\leftarrow b_{j, k} / \sum_kb_{j, k}$ and rescale $\mathcal{H}_2$ and $\mathcal{U}_2$ accordingly.

\subsection{ILRMA}
ILRMA is a special class of MNMF designed to solve determined source separation problems.
Unlike MNMF, which uses the mixing system (\eqcite{mixing}), ILRMA uses a separation system of the form
\begin{align}
	\label{eq:separation}
	\bm{s}(f, n) &= \bm{W}^\sf{H}(f)\bm{x}(f, n), \\
	\bm{W}(f) &= \left[ \bm{w}_1(f), \cdots, \bm{w}_I(f) \right] \in \mathbb{C}^{I \times J},
\end{align}
assuming the mixing matrix is invertible.
The inverse matrix $\bm{W}^\sf{H}(f)$ is called the separation matrix.
From \eqcite{LGMs} and \eqcite{separation}, $\bm{x}(f, n)$ is shown to follow
\begin{align}
	\label{eq:observed2}
	\bm{x}(f, n) \sim \Nc{\bm{x}(f, n)}{\bm{0}}{(\bm{W}^\sf{H}(f))^{-1}\bm{V}(f, n)(\bm{W}(f))^{-1}}.
\end{align}
The log-likelihood of the separation matrices $\mathcal{W}=\left\{ \bm{W}(f) \right\}_f$ and $\mathcal{V}$ given the observed signals $\mathcal{X}$ is given by
\begin{align}
	\label{eq:likelihood2}
	&\log p(\mathcal{X} | \mathcal{W}, \mathcal{V}) \nonumber \\
	&\overset{c}{=} 2N \sum_f \log |\rm{det} \bm{W}^\sf{H}(f)|\nonumber \\
	&\quad- \sum_{f, n, j}\left[ \log v_j(f, n) + \frac{ |\bm{w}_j^\sf{H}(f)\bm{x}(f, n)|^2 }{ v_j(f, n) } \right],
\end{align}
where $v_j(f, n)$ is modeled as \eqcite{nmf} or  \eqcite{nmf2} as with MNMF.

As with MNMF, we can derive MM-based update equations for $\mathcal{H}_1$, $\mathcal{U}_1$ or $\mathcal{B}$, $\mathcal{H}_2$, $\mathcal{U}_2$.
Since ILRMA is a natural extension of IVA, we can use a fast update rule called the Iterative Projection (IP)~\cite{ono2011stable} for the separation matrices, originally developed for IVA.

\subsection{DNN Approach}
There has been some studies attempting to integrate deep neural networks (DNNs) with the LGM-based multichannel source separation framework~\cite{nugraha2016multichannel}.
\cite{nugraha2016multichannel} proposes an algorithm that consists of updating $v_j(f, n)$ for each $j$ via the forward computation of a pretrained DNN.
Here, each DNN is trained so that it produces a denoised version of the input spectra.
Hence, $v_j(f, n)$ is forced to get close to the spectra of clean speech at each iteration.
However, updating $v_j(f, n)$ in this way does not guarantee an increase in the log-likelihood, which does not ensure the convergence of the devised algorithm.

\subsection{MVAE}
One limitation of the MNMF framework including ILRMA is that since $v_j(f, n)$ is restricted to \eqcite{nmf}, it can fail to work for sources with spectrograms that do not actually follow this form.
The MVAE method is an extension of ILRMA that replaces \eqcite{nmf} with a pretrained CVAE.
Let $\tilde{\bm{S}}=\left\{ s(f, n) \right\}_{f, n}$ be the complex spectrogram of a particular sound source.
MVAE models the generative model of $\tilde{\bm{S}}$ using a Conditional VAE (CVAE) with an auxiliary input $c$.
Here, we assume that $c$ is represented as a one-hot vector, indicating the class of a source.
Thus, the elements of $c$ must sum to unity.
For example, if we consider speaker identities as the source class, each element of $c$ will be associated with a different speaker.

The CVAE consists of an encoder network and a decoder network, which are assumed to be trained using labeled training examples $\{ \tilde{\bm{S}}_m, c_m \}_{m=1}^M$ prior to separation.
The encoder distribution $q_\phi(\bm{z}|\tilde{\bm{s}}, c)$ is expressed as a Gaussian distribution:
\begin{align}
	\label{eq:q}
	q_\phi(\bm{z}|\tilde{\bm{S}}, c) = \prod_k\N{z(k)}{\mu_\phi(k;\tilde{\bm{S}}}{\sigma_\phi^2(k;\tilde{\bm{S}}, c)},
\end{align}
where $\bm{z}$ denotes a latent space variable and $z(k)$, $\mu_\phi(k;\tilde{\bm{S}})$, and $\sigma_\phi^2(k;\tilde{\bm{S}}, c)$ represent the $k$--th elements of $\bm{z}$, $\bm{\mu}_\phi(\tilde{\bm{S}}, c)$, and $\bm{\sigma}_\phi^2(\tilde{\bm{S}}, c)$, respectively.
The decoder distribution $p_\theta(\tilde{\bm{S}}|\bm{z}, c, g)$ is expressed as a zero-mean complex Gaussian distribution (LGM):
\begin{align}
	\label{eq:p}
	p_\theta(\tilde{\bm{S}}|\bm{z}, c, g) &= \prod_{f, n}\Nc{s(f, n)}{ 0 }{ v(f, n) }, \\
	v(f, n) &= g \cdot \sigma_\theta^2(f, n; \bm{z}, c),
\end{align}
where $\sigma_\theta^2(f, n; \bm{z}, c)$ represents the $(f, n)$--th elements of the decoder output  $\bm{\sigma}_\theta^2(\bm{z}, c)$ and $g$ is the global scale of the generated spectrogram.
Both the encoder and decoder network parameters $\phi$, $\theta$ are trained using the following objective fucntion
\begin{align}
	\label{eq:VAE}
	\mathcal{J}(\phi, \theta) = \mathbb{E}_{ (\tilde{\bm{S}}, c) \sim p_\rm{D} (\tilde{\bm{S}}, c) } \left[ \mathbb{E}_{ \bm{z} \sim q(\bm{z} | \tilde{\bm{S}}, c) } [ \log p(\tilde{\bm{S}}|\bm{z}, c) ] \right. \nonumber \\
	\left.- \rm{KL} [ q(\bm{z}|\tilde{\bm{S}}, c) ||p(\bm{z}) ] \right],
\end{align}
where $\mathbb{E}_{ (\tilde{\bm{S}}, c) \sim p_\rm{D} (\tilde{\bm{S}}, c) }[\cdot]$ denotes the sample mean over the training examples and $\rm{KL} [ \cdot ||\cdot ] $ is the Kullback-Leibler divergence.

The trained decoder distribution $p_\theta(\tilde{\bm{S}}|\bm{z}, c, g)$ is considered as a universal generative model that is capable of generating spectrograms of all the sources involved in the training examples.
MVAE employs the decoder part of the CVAE as the source model $v_j(f, n)$ in \eqcite{likelihood2} and treats the input $\bm{z}$ and $c$ to the decoder as the model parameters to be estimated.
The optimization algorithm of MVAE consists of updating the separation matrices using IP, the global scale and the input to the pretrained decoder using backpropagation.
The advantage of the MVAE is that it can leverage the strong representation power of VAE for source power spectrogram modeling.

\begin{figure}[t]
	\centering
	\includegraphics[width=\columnwidth]{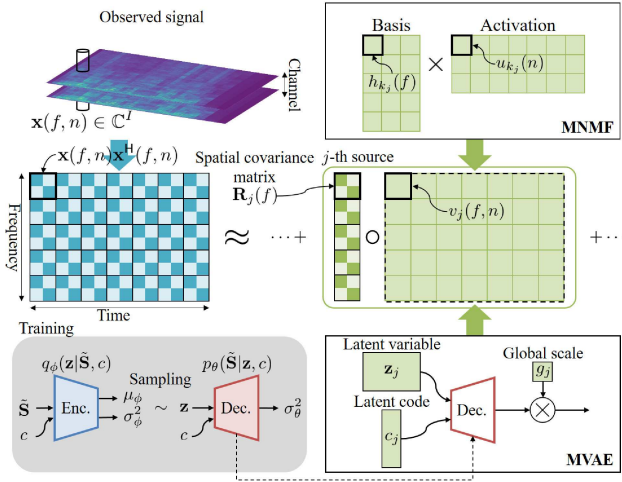}
	\vspace{-9pt}
	\caption{Illustration of Generalized MVAE}
	\vspace{-9pt}
	\label{fig:overview}
\end{figure}
\section{GENERALIZED MVAE}
While the MVAE method is applicable only to determined mixtures, we propose generalizing it to so that it can also deal with underdetermined mixtures.
As with the original MVAE method, we use the decoder network of the pretrained CVAE as the generative model of source power spectrograms.
\figcite{overview} shows an illustration of GMVAE and MNMF with the source model given by \eqcite{nmf}.

Since the decoder distribution is given in the same form as the LGM, we can use $p_\theta(\tilde{\bm{S}}_j | \bm{z}_j, c_j, g_j)$ to develop the log-likelihood of the form \eqcite{likelihood}.
Hence, we can derive an iterative algorithm for estimating $\mathcal{R}$, $\mathcal{G}=\left\{ g_j \right\}_j$, and $\Psi=\left\{ \bm{z}_j, c_j \right\}_j$ in the same way as the derivation of the MM-based algorithm for MNMF.
According to~\cite{kameoka2018general}, we can show that
\begin{align}
	\mathcal{L} &= -\log p(\mathcal{X} | \mathcal{A}, \mathcal{V}) \nonumber \\
	&\overset{c}{\le}\sum_j\sum_{f, n}\biggl[ \frac{ \tr{(\sf{X}(f, n) \bm{P}_j(f, n) \bm{R}_j^{-1}(f, n) \bm{P}_j(f, n) )} }{ v_j(f, n) } \nonumber \\
	&\quad \qquad \qquad +v_j(f, n)\tr(\bm{K}^{-1}(f, n)\bm{R}_j(f, n)) \biggr],
\end{align}
where the equality holds when
\begin{align}
	\label{eq:P}
	\bm{P}_j(f, n) &= v_j(f, n)\bm{R}_j(f, n)\left( \sum_j v_j(f, n)\bm{R}_j(f, n) \right)^{-1}, \\
	\label{eq:K}
	\bm{K}(f, n) &= \sf{X}(f, n).
\end{align} 
Thus, we can use the right-hand side of this inequality as a majorizer of $\mathcal{L}$ where $\mathcal{P}=\left\{\bm{P}_j(f, n)\right\}_{j, f, n}$ and $\mathcal{K}=\left\{\bm{K}(f, n)\right\}_{f, n}$ are auxiliary variables.
An iterative algorithm that consists of minimizing this majorizer with respect to $\mathcal{R}$, $\mathcal{G}$, and $\Psi$ and updating $\mathcal{P}$ and $\mathcal{K}$ at \eqcite{P} and \eqcite{K} is guaranteed to converge to a stationary point of $\mathcal{L}$.
The optimal update of $\mathcal{R}$ is given as the solution to \eqcite{ricatti}.
Since the majorizer is split into source-wise terms, $\Psi$ can be updated parallelly using backpropagation.
Note that we must take account of the sum-to-one constraints when updating $c_j$.
This can be easily implemented by inserting an appropriately designed softmax layer that outputs $c_j$
\begin{align}
	c_j = \rm{softmax}(d_j),
\end{align}
and treating $d_j$ as the parameter to be estimated instead.
The optimal update of $\mathcal{G}$ is obtained as
\begin{align}
	\label{eq:g}
	&g_j \leftarrow g_j \nonumber \\
	&\times \sqrt{ \frac{ \sum_{f, n} \sigma_\theta^2(f, n; \bm{z}_j, c_j) \rm{tr}(\hat{\sf{X}}^{-1}(f, n)\sf{X}(f, n)\hat{\sf{X}}^{-1}(f, n)\bm{R}_j(f)) }{ \sum_{f, n} \sigma_\theta^2(f, n; \bm{z}_j, c_j) \rm{tr}(\hat{\sf{X}}^{-1}(f, n)\bm{R}_j(f)) } }.
\end{align}
The source separation algorithm of GMVAE is summarized as follows:
\begin{enumerate}
	\item Train $\phi$ and $\theta$ using \eqcite{VAE}.
	\item Initialize $\mathcal{R}$, $\mathcal{G}$, and $\Psi = \left\{ \bm{z}_j, c_j \right\}$.
	\item Iterate the following steps for each $j$:
	\begin{enumerate}
		\item Update $\mathcal{R}$ using \eqcite{ricatti}, \eqcite{ricatti_1}, and \eqcite{ricatti_2}.
		\item Update $\psi_j = \left\{ \bm{z}_j, c_j \right\}$ using backpropagation.
		\item Update $g_j$ using \eqcite{g}.
	\end{enumerate}
\end{enumerate}

\section{Experimental Evaluations}
\begin{figure}[t]
	\centering
	\includegraphics[width=\columnwidth]{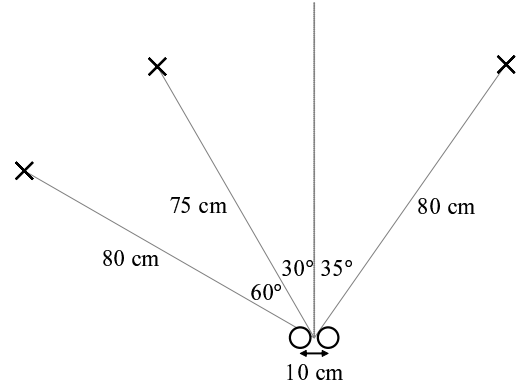}
	\vspace{-9pt}
	\caption{Microphone and source positions}
	\label{fig:room}
\end{figure}
\begin{figure*}[t]
	\begin{tabular}{cccc}
		\begin{minipage}{.24\hsize}
			\centering
			\includegraphics[width=\columnwidth]{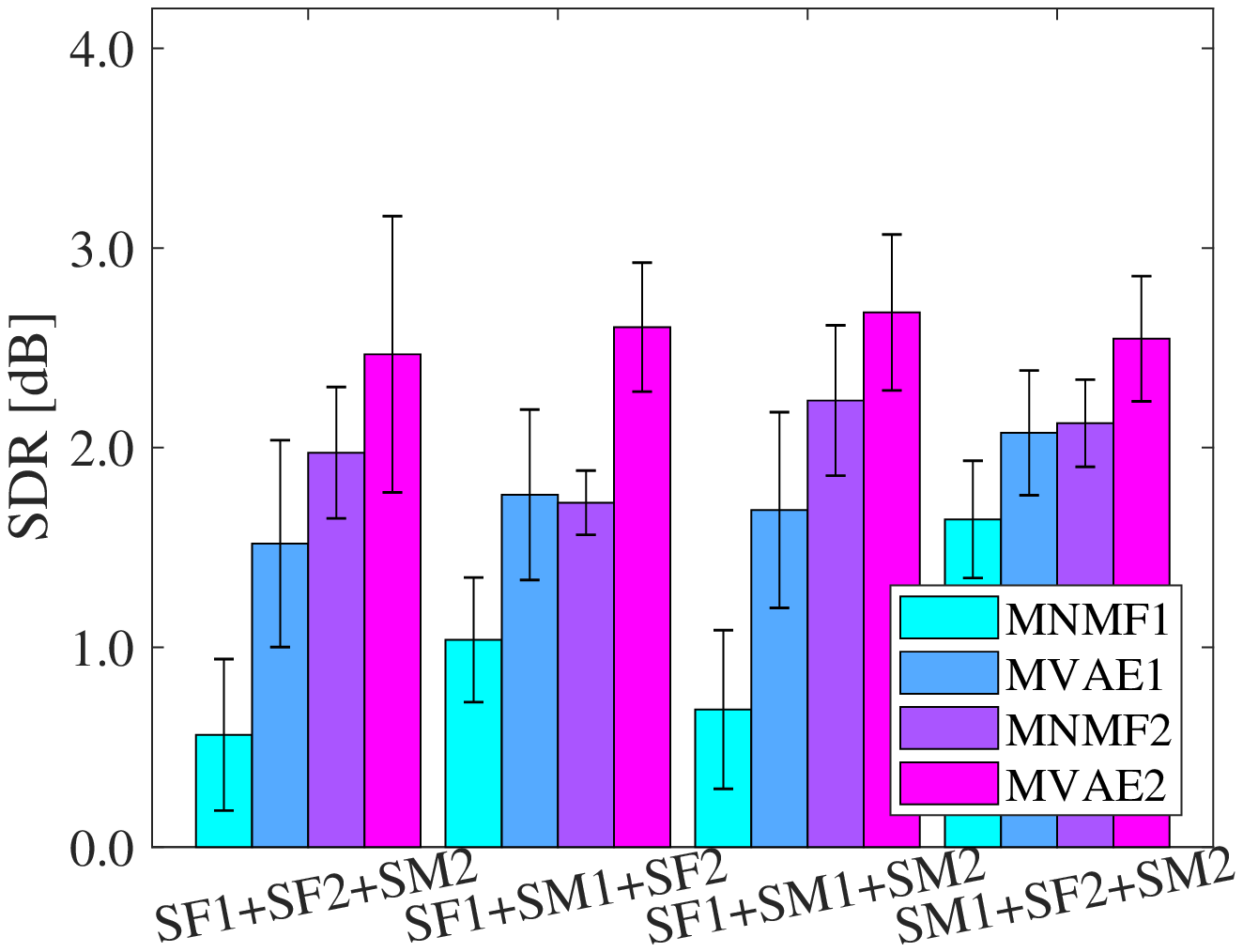}
		\end{minipage}
		\begin{minipage}{.24\hsize}
			\centering
			\includegraphics[width=\columnwidth]{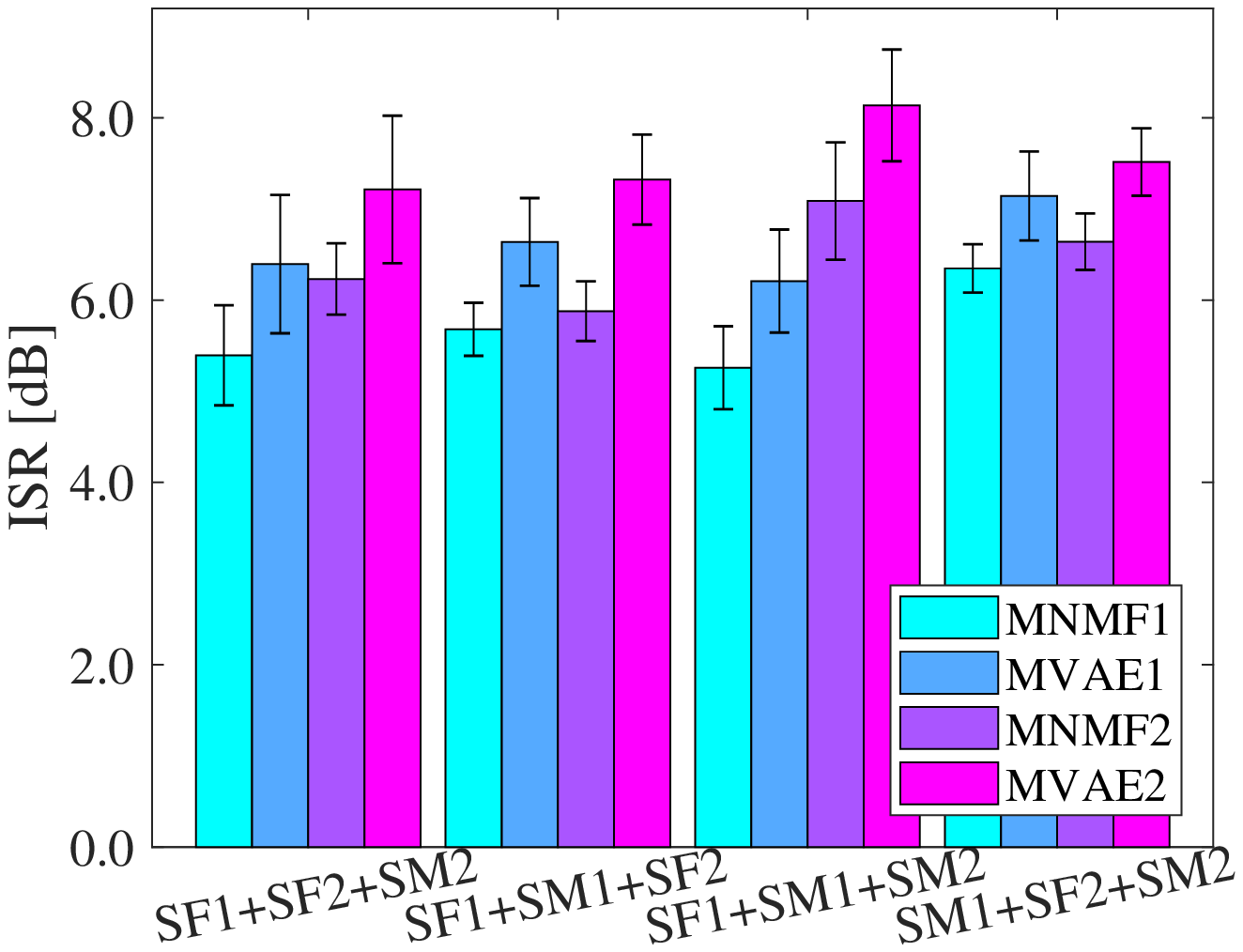}
		\end{minipage}
		\begin{minipage}{.24\hsize}
			\centering
			\includegraphics[width=\columnwidth]{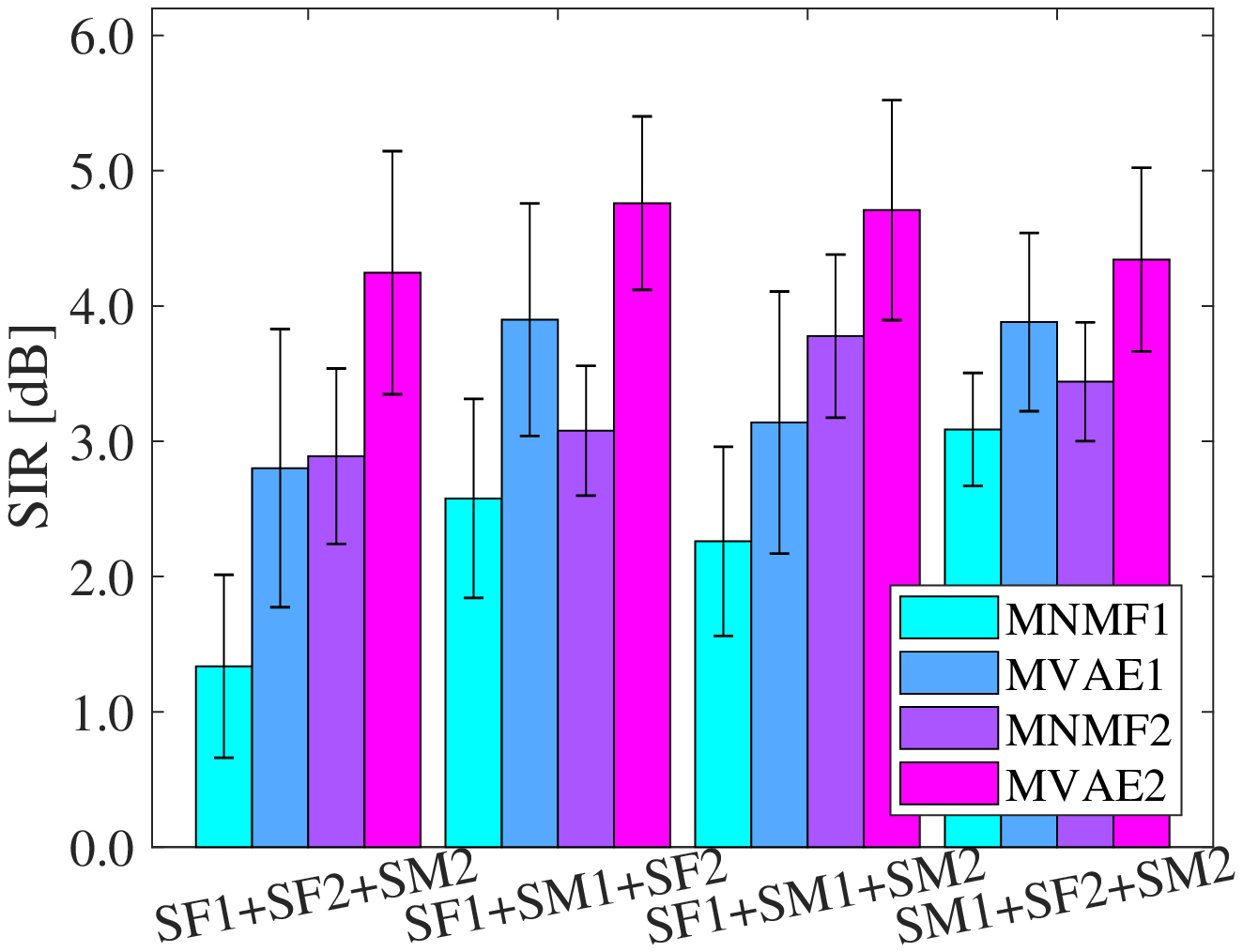}
		\end{minipage}
		\begin{minipage}{.24\hsize}
			\centering
			\includegraphics[width=\columnwidth]{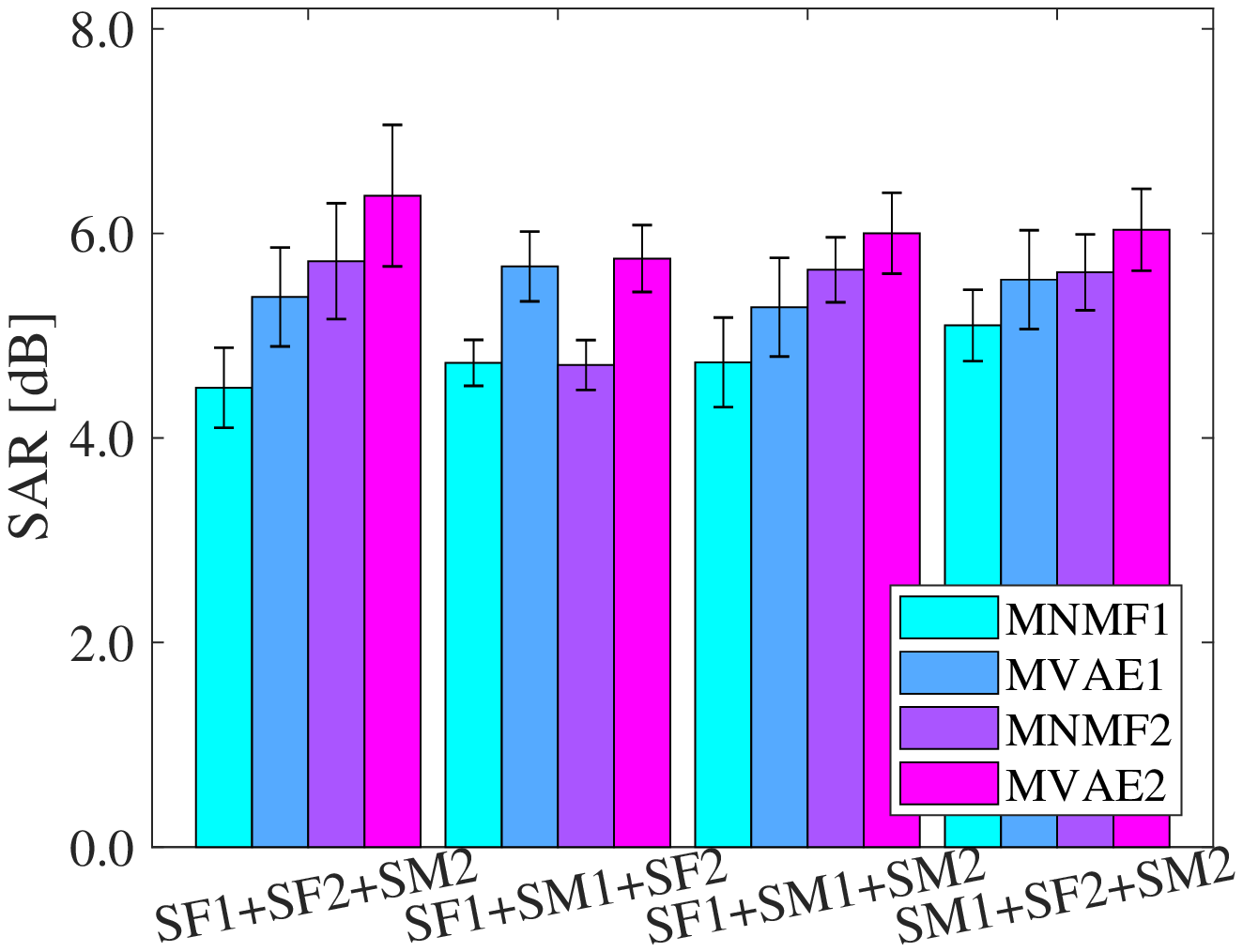}
		\end{minipage}
	\end{tabular}
	\vspace{-9pt}
	\caption{Averaged separation performances for $T_{60} = 78~[\rm{ms}]$}
	\label{fig:78ms}
\end{figure*}
\begin{figure*}[t]
	\begin{tabular}{cccc}
		\begin{minipage}{.24\hsize}
			\centering
			\includegraphics[width=\columnwidth]{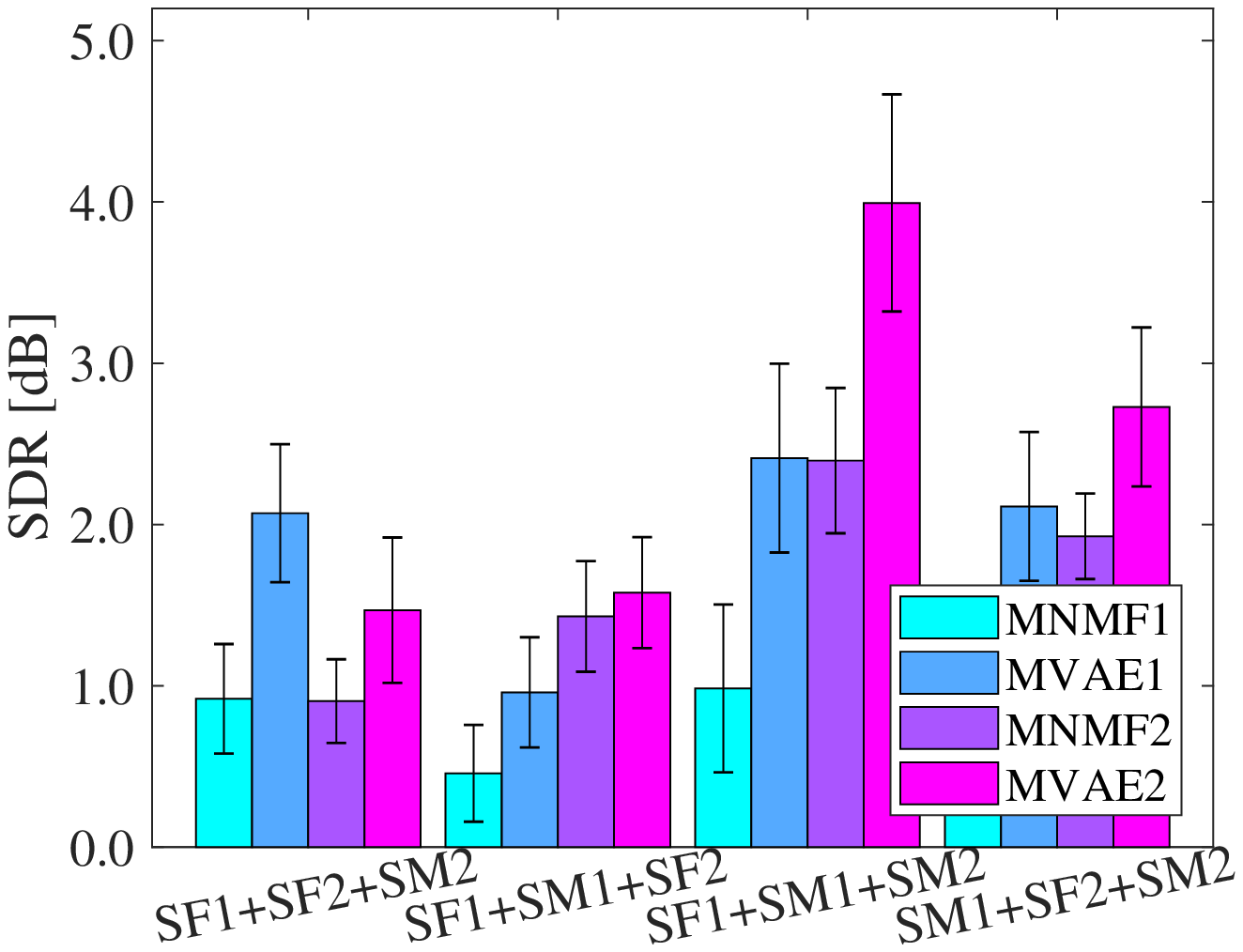}
		\end{minipage}
		\begin{minipage}{.24\hsize}
			\centering
			\includegraphics[width=\columnwidth]{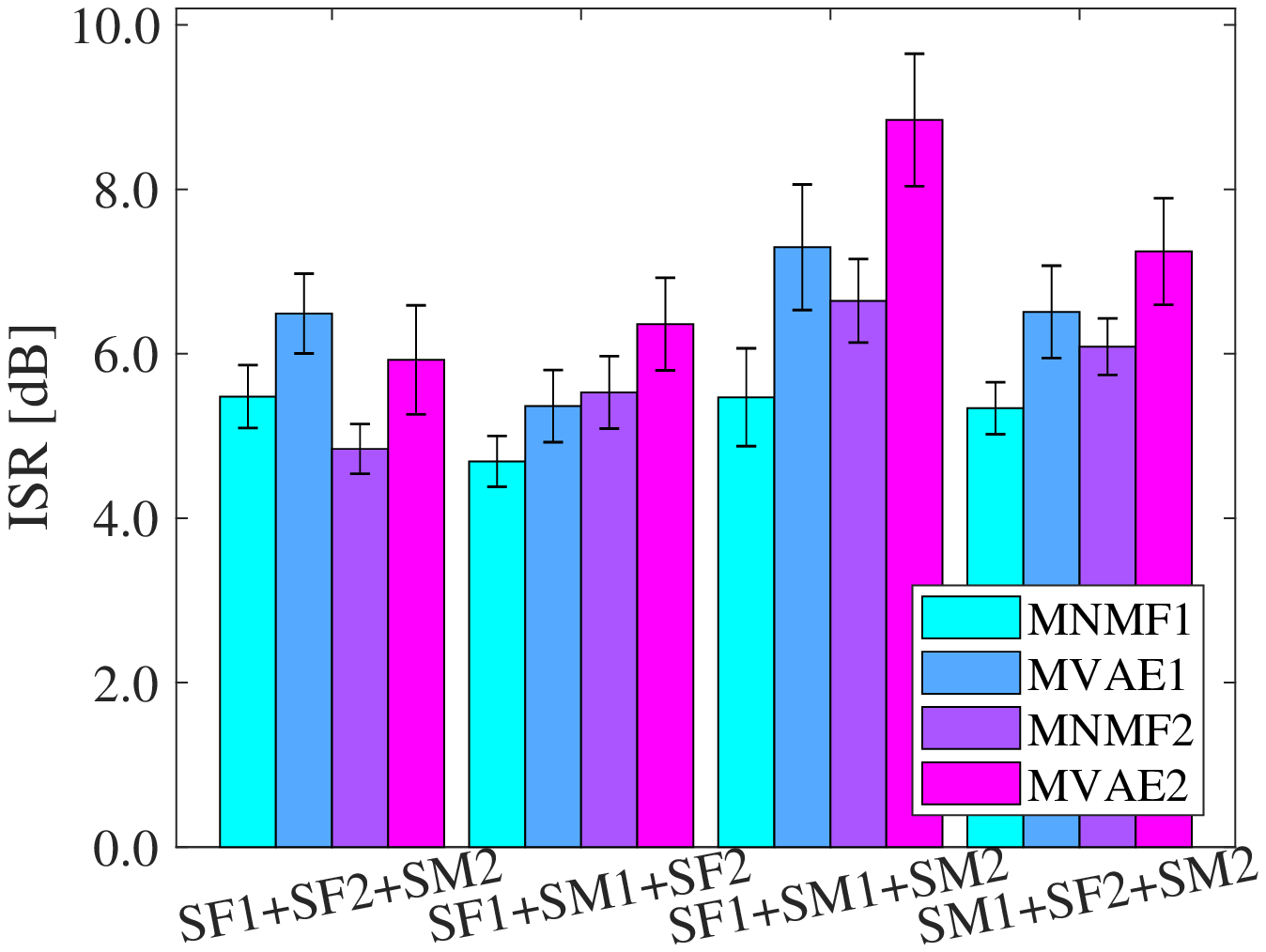}
		\end{minipage}
		\begin{minipage}{.24\hsize}
			\centering
			\includegraphics[width=\columnwidth]{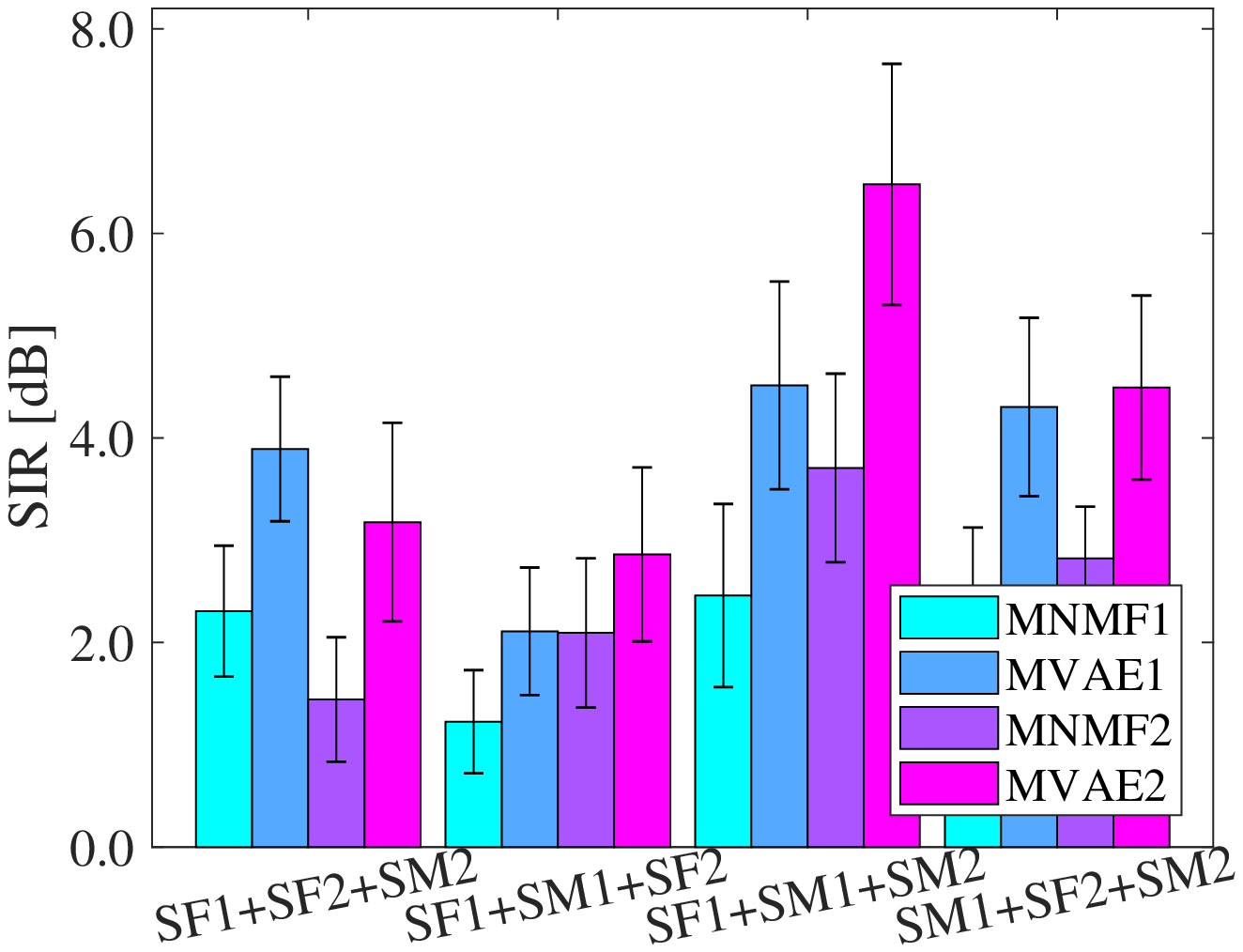}
		\end{minipage}
		\begin{minipage}{.24\hsize}
			\centering
			\includegraphics[width=\columnwidth]{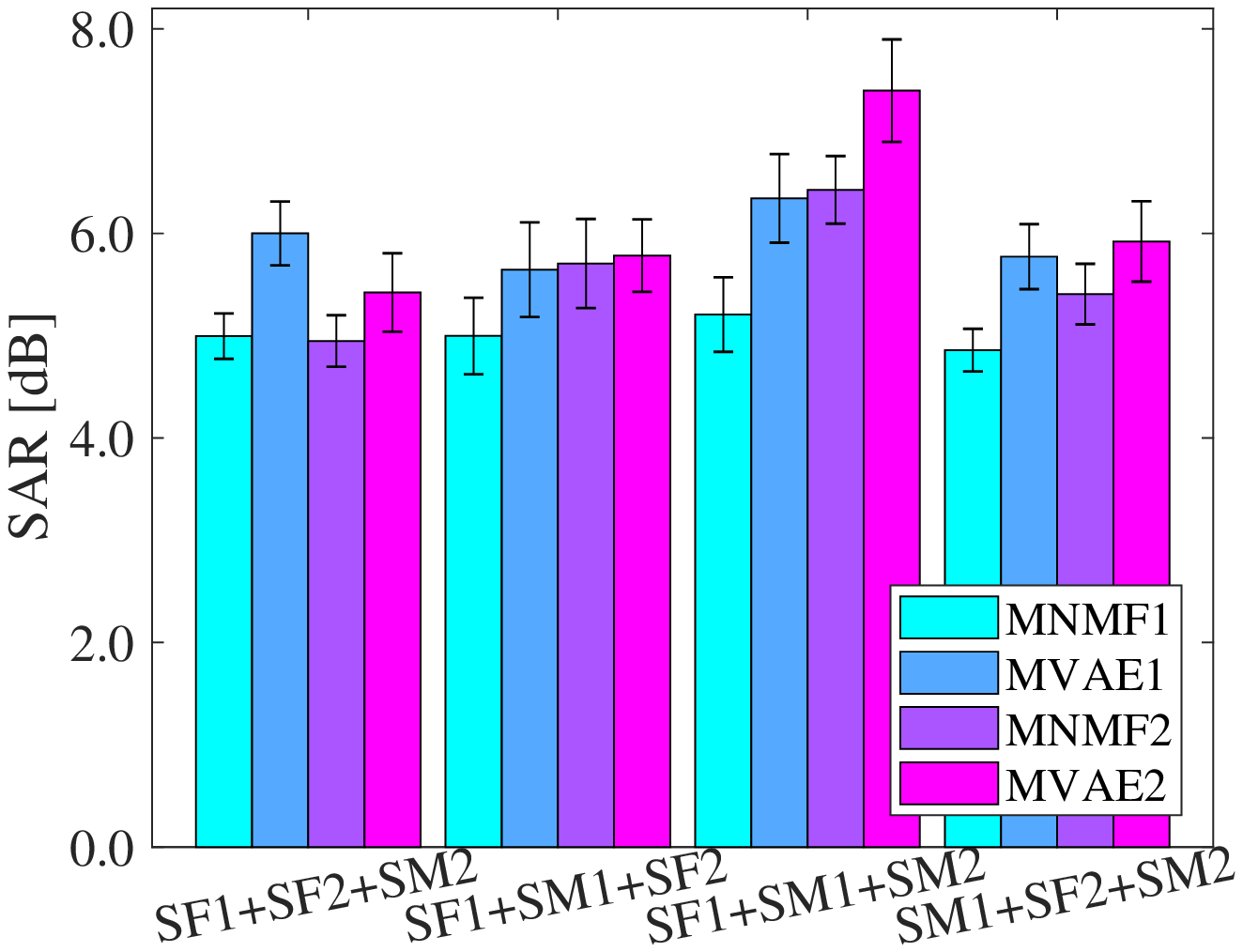}
		\end{minipage}
	\end{tabular}
	\vspace{-9pt}
	\caption{Averaged separation performances for $T_{60} = 351~[\rm{ms}]$}
	\label{fig:351ms}
\end{figure*}
The proposed method was evaluated on an underdetermined source separation task of separating out three sources from two microphone inputs.
As the experimental data, we used speech samples of the Voice Conversion Challenge (VCC) 2018 dataset~\cite{lorenzo2018voice}, which contains recordings of six female and six male US English speakers.
Specifically, we used a set of the utterances of two female and two male speakers, 'SF1', 'SF2', 'SM1', and 'SM2'.
In this experiment, speaker identities are considered as the source class category: $c$ is represented as a four-dimensional one-hot vector.
81 sentences and 35 sentences of each speaker were used for training and evaluation, respectively.
\figcite{room} shows the position of microphones and sources.
$\Circle$ and $\times$ show the microphones and sources, respectively.
10 speech mixtures are generated for four speaker patterns: SF1+SF2+SM2, SF1+SM1+SF2, SF1+SM1+SM2, and SM1+SF2+SM2.

All the speech signals were resampled at 16~[kHz] and STFT analysis was conducted with 256~[ms] frame length and 128~[ms] hop length.
We designed the encoder and decoder networks of the CVAE as a three-layer fully-convolutional network with gated linear units and a three-layer fully-deconvolutional network with gated linear units as in~\cite{kameoka2018semi}.
The Adam~\cite{kingma2015adam} algorithm with learning rate 0.0002 was used to train the CVAE and the Stochastic Gradient Descent (SGD) algorithm with learning rate 0.0005 was used to update the VAE source model $\Psi$.
We chose the MNMF algorithm with the source model given by \eqcite{nmf} or \eqcite{nmf2} as baseline methods (MNMF1, MNMF2) for comparison.
The separation algorithm was run for 300 iterations for the conventional methods and 100 iterations for the proposed.
The parameters of the proposed method were initialized using the baseline method run for 200 iterations.
Therefore, we tested the proposed method with two different initial settings (MVAE1, MVAE2).
As the evaluation metrics, the Signal-to-Distortion Ratio (SDR), the source Image-to-Spatial distortion ratio (ISR), the Signal-to-Inference Ratio (SIR), and the Signal-to-Artifact Ratio (SAR)~\cite{vincent2006performance} between the reference signals and the separated signals were calculated for each mixture and averaged over 10 samples in each speaker pattern.
Separation performance was investigated with two different reverberant conditions where the reverberation times $T_{60}$ were set to 78~[ms] and 351~[ms], respectively.

The separation performance under each reverberant condition is shown in \figcite{78ms} and \ref{fig:351ms}.
We can see that the proposed method obtained better separation performance than the baseline methods.
The results imply that the use of VAE source model has successfully contributed to improving the separation performance.

\section{Conclusion}
	This paper proposed the GMVAE method, which generalizes the MVAE method so that it can also be applied to multichannel source separation method under underdetermined conditions.
	Experimental results revealed that the GMVAE method achieved better performance than the baseline method.


\bibliographystyle{IEEEbib}
\bibliography{arXivseki}

\end{document}